\documentclass[runningheads]{llncs}

\usepackage{amsmath}
\usepackage{amssymb}

\usepackage{siunitx}

\usepackage{multirow}

\usepackage{hyperref}
\usepackage{xcolor}

\usepackage{booktabs}
\usepackage{array}

\newcolumntype{x}[1]{%
>{\centering\hspace{0pt}}p{#1}}%

\usepackage{doc}
\usepackage{graphicx}

\newcommand*{\M}{\mathcal{M}}
\newcommand*{\W}{\mathcal{W}}
\renewcommand*{\S}{\mathcal{S}}
\renewcommand*{\P}{\mathcal{P}}
\newcommand*{\T}{\mathcal{T}}
\newcommand{\R}{\mathbf{R}}
\newcommand{\Cl}{\mathit{Cl}}
\newcommand{\Conj}{\mathit{Conj}}
\newcommand{\Fin}{\mathit{Fin}}

\newcommand*{\Mlin}{\M_\textrm{lin}}
\newcommand*{\Mxgb}{\M_\textrm{tree}}
\newcommand*{\Mnn}{\M_\textrm{nn}}

\title{ENIGMA-NG: Efficient Neural and Gradient-Boosted Inference Guidance for E%
\thanks{Supported by the ERC
Consolidator grant no. 649043 AI4REASON, and by the Czech project AI\&Reasoning 
CZ.02.1.01/0.0/0.0/15\_003/0000466 and the European Regional Development Fund.}}

\author{
   Karel Chvalovsk\'{y} \and Jan Jakub\r{u}v \and Martin Suda \and Josef Urban
}

\institute{
  Czech Technical University in Prague, Prague,
  Czech Republic\\
 }

\authorrunning{Chvalovsk\'{y} et al.}

\titlerunning{Enhancing ENIGMA Given Clause Predictions with Conjecture
Features}

\makeatletter
\renewcommand\section{\@startsection{section}{1}{\z@}%
                       {-12\p@ \@plus -4\p@ \@minus -4\p@}%
                       {8\p@ \@plus 4\p@ \@minus 4\p@}%
                       {\normalfont\large\bfseries\boldmath
                        \rightskip=\z@ \@plus 8em\pretolerance=10000 }}
\makeatother

\begin{document}

\maketitle

\begin{abstract}
  We describe an efficient implementation of clause guidance in
  saturation-based automated theorem provers extending the ENIGMA
  approach. Unlike in the first ENIGMA implementation where fast
  linear classifier is trained and used together with manually
  engineered features, we have started to experiment with more
  sophisticated state-of-the-art machine learning methods such as
  gradient boosted trees and recursive neural networks. In particular
  the latter approach poses challenges in terms of efficiency of
  clause evaluation, however, we show that deep integration of the
  neural evaluation with the ATP data-structures can largely amortize
  this cost and lead to competitive real-time results.  Both methods
  are evaluated on a large dataset of theorem proving problems and
  compared with the previous approaches.  The resulting methods
  improve on the manually designed clause guidance, providing the
  first practically convincing application of gradient-boosted and
  neural clause guidance in saturation-style automated theorem provers.

\end{abstract}

\section{Introduction}

Automated theorem provers (ATPs)~\cite{DBLP:books/el/RobinsonV01} have
been developed for decades by manually designing proof calculi and
search heuristics.  Their power has been growing and they are already
very useful, e.g., as parts of large interactive theorem proving (ITP)
verification toolchains (hammers)~\cite{hammers4qed}.  On the other
hand, with small exceptions, ATPs are still significantly weaker than
trained mathematicians in finding proofs in most research domains.

Recently, machine learning over large formal corpora created from ITP libraries~\cite{Urban06,MengP08,holyhammer}
 has started to be used to develop
guidance of ATP systems~\cite{US+08-long,KuhlweinLTUH12-long,IrvingSAECU16}. 
This has already produced strong systems for selecting
relevant facts for proving new conjectures over large formal
libraries~\cite{abs-1108-3446,BlanchetteGKKU16,hh4h4}. More
recently, machine learning has also  started to be used to guide the internal search of the ATP systems. In
sophisticated saturation-style provers this has been done by feedback
loops for strategy invention~\cite{blistr,JakubuvU18a,SchaferS15} and
by using supervised learning~\cite{JakubuvU17a,LoosISK17} to select
the next given clause~\cite{mccune1990otter}. In the simpler connection
tableau systems such as leancop~\cite{OB03}, supervised learning has
been used to choose the next tableau extension
step~\cite{UrbanVS11,KaliszykU15} and first experiments with
Monte-Carlo guided proof search~\cite{FarberKU17} and reinforcement learning~\cite{KaliszykUMO18} 
have been done.

In this work, we add two state-of-the-art machine learning methods to
the ENIGMA~\cite{JakubuvU17a,JakubuvU18} algorithm that efficiently guides saturation-style proof
search. The first one trains gradient boosted trees on efficiently
extracted manually designed (handcrafted) clause features. The second method removes
the need for manually designed features, and instead uses end-to-end
training of recursive neural networks. Such architectures, when implemented naively,
are typically expensive and may be impractical for saturation-style ATP. We show that 
 deep integration of the neural evaluation with the ATP data-structures can largely amortize
  this cost, allowing competitive performance.

The rest of the paper is structured as follows.
Section~\ref{sec:guidance} introduces the saturation-based automated theorem
proving with the emphasis on machine learning.
Section~\ref{sec:hand} briefly summarizes our previous work with handcrafted
features in ENIGMA, it extends previously published ENIGMA with additional
classifier based on decision trees (Section~\ref{sec:decision}) and simple
feature hashing (Section~\ref{sec:hashing}).
Section~\ref{sect:neural_arch} presents our new approach to apply neural
networks for ATP guidance.
Section~\ref{sec:experiments} provides experimental evaluation of our work and
we conclude in Section~\ref{sec:conclude}.

\section{Automated Theorem Proving with Machine Learning}
\label{sec:guidance}

State-of-the-art saturation-based automated theorem provers (ATPs) for
first-order logic (FOL), such as E \cite{Sch02-AICOMM} and Vampire~\cite{Vampire}
are today's most advanced tools for general reasoning across a variety of
mathematical and scientific domains. 
Many ATPs employ the \emph{given clause algorithm}, translating
the input FOL problem $T\cup\{\lnot C\}$ into a refutationally
equivalent set of clauses.
The search for a contradiction is performed maintaining sets of
\emph{processed} ($P$) and \emph{unprocessed} ($U$) clauses.
The algorithm repeatedly selects a \emph{given clause} $g$ from $U$,
moves $g$ to $P$, and extends $U$ with all clauses inferred with $g$ and $P$.
This process continues until a contradiction is found, $U$ becomes empty, or 
a resource limit is reached.
The search space of this loop grows quickly and it is a well-known fact that the
selection of the right given clause is crucial for success.
Machine learning from a large number of proofs and proof searches may help guide the selection of the given clauses.

E allows the user to select a \emph{proof search strategy} $\S$ to guide the
proof search.
An E strategy $\S$ specifies parameters such as term ordering, literal selection
function, clause splitting, paramodulation setting, premise selection, and, most
importantly for us, the \emph{given clause selection} mechanism.
The given clause selection in E is implemented using a list of priority queues.
Each priority queue stores all the generated clauses in a specific order
determined by a clause \emph{weight function}.
The clause weight function assigns a numeric (real) value to each clause, and
the clauses with smaller weights (``lighter clauses'') are prioritized.
To select a given clause, one of the queues is chosen in a round robin manner,
and the clause at the front of the chosen queue gets processed.
Each queue is additionally assigned a \emph{frequency} which amounts to the
relative number of clause selections from that particular queue.
Frequencies can be used to prefer one queue over another.
We use the following notation to denote the list of priority queues with
frequencies $f_i$ and weight functions $\W_i$:
\[
   (f_1*\W_1,\ldots,f_k*\W_k).
\]

To facilitate machine learning research, E implements an option
under which each successful proof search gets analyzed and 
the prover outputs a list of clauses annotated as either \emph{positive} or \emph{negative} training examples.
Each processed clause which is present in the final proof is classified as
positive. %
On the other hand, processing of clauses not present in the final proof was redundant,
hence they are classified as negative.
Our goal is to learn such classification (possibly conditioned on the
problem and its features) in a way that generalizes and allows solving
related problems.

Given a set of problems $\P$, we can run E with a strategy $\S$ and obtain positive and
negative training data
$\T$ from each of the successful proof searches.
In this work, we use three different machine learning methods to learn the clause
classification given by $\T$, each method yielding a \emph{classifier} or \emph{model} $\M$.
Each of the machine learning methods has a different structure of $\M$ described
in detail later.
With any method, however, $\M$ provides the function to compute the weight
of an arbitrary clause.
This weight function is then used in E to guide further proof runs.

A model $\M$ can be used in E in different ways.
We use two methods to combine $\M$ with a strategy $\S$.
Either (1) we use $\M$ to select \emph{all} the given clauses, or (2) we combine $\M$
with the given clause guidance from $\S$ so that roughly half of the clauses are
selected by $\M$.
We denote the resulting E strategies as (1) $\S\odot\M$, and (2) $\S\oplus\M$.
The two strategies are equal up to the priority queues for given clause
selection which are changed ($\rightsquigarrow$) as follows.
\[\begin{array}{lllll}
   \mbox{in }\S\odot\M: &\quad& (f_1*\W_1,\ldots,f_k*\W_k) &\rightsquigarrow& (1*\M), \\
   \mbox{in }\S\oplus\M: &\quad& (f_1*\W_1,\ldots,f_k*\W_k) &\rightsquigarrow&
   ((\sum{f_i})*\M, f_1*\W_1, \ldots, f_k*\W_k).
\end{array}\]

The strategy $\S\oplus\M$ usually performs better in practice as it helps to counter
overfitting by combining powers with the original strategy $\S$.
The strategy $\S\odot\M$ usually provides additional proved problems, gaining additional
training data, and it is useful for the evaluation of the training phase.
When $\S\odot\M$ performs better than $\S$, it indicates that $\M$ has learned
the training data well.
When it performs much worse, it indicates that $\M$ is not very well
trained.
The strategy $\S\oplus\M$ should always perform better than $\S$, otherwise the
guidance of $\M$ is not very useful.
Additional indication of successful training can be obtained from
the number of clauses processed during a successful proof search.
The strategy $\S\odot\M$ should run with much less processed clauses, in some cases
even better than $\S\oplus\M$ as the original $\S$ might divert the proof
search.
In the best case, when $\M$ learns some problem perfectly, the number of
processed clauses is approaching the length of the proof.

It is important, however, to combine a model $\M$ only with a ``compatible''
strategy $\S$.
For example, let us consider a model $\M$ trained on samples obtained with
another strategy $S_0$ which has a different term ordering than $\S$.
As the term ordering can change term normal forms, the clauses
encountered in the proof search with $\S$ might look quite different than the
training clauses.
This causes problems unless the trained models are independent of symbol names, which is
not (yet) our case.
Additional problems might be caused as term orderings and literal selection
might change the proof space and the original proofs might not be reachable.
Hence we only combine $\M$ with the strategy $\S$ which provided the examples 
on which $\M$ was trained.

\section{ATP Guidance with Handcrafted Clause Features}
\label{sec:hand}

In order to employ a machine learning method for ATP guidance, first-order
clauses need to be represented in a format recognized by the selected learning
method.
A common approach is to manually extract a finite set of various properties of
clauses called \emph{features}, and to encode these clause features by a 
fixed-length numeric vector.
Various machine learning methods can handle numeric vectors and their success
heavily depends on the selection of correct clause features.
In this section, we work with handcrafted clause features which, we believe,
capture the information important for ATP guidance.

ENIGMA~\cite{JakubuvU17a,JakubuvU18} is our \emph{efficient}
learning-based method for guiding given clause selection in saturation-based
ATPs.
Sections~\ref{sec:enigma} and~\ref{sec:regression} briefly summarizes our
previous work.
Sections~\ref{sec:decision} and~\ref{sec:hashing} describe extensions
first-presented in this work.

\subsection{ENIGMA Clause Features}
\label{sec:enigma}

So far, the development of ENIGMA was focusing on fast and practically usable methods,
allowing E users to directly benefit from our work. 
Various possible choices of efficient clause features for theorem prover
guidance have been experimented
with~\cite{JakubuvU17a,JakubuvU18,KaliszykUMO18,DBLP:conf/ijcai/KaliszykUV15}.
The original ENIGMA~\cite{JakubuvU17a} uses term-tree walks of
length 3 as features, while the second version~\cite{JakubuvU18} reaches better
results by employing various additional features.
In particular, the following types of features are used 
(see~\cite[Sec.~3.2]{JakubuvU17a} and~\cite[Sec.2]{JakubuvU18} for
details):

\begin{description}
\item[Vertical Features] are (top-down-)oriented term-tree walks of length
3.  For example, the unit clause $P(f(a,b))$ contains only
features $(P,f,a)$ and $(P,f,b)$.
\item[Horizontal Features] are horizontal cuts of a term tree.
For every term $f(t_1,\ldots,t_n)$ in the clause, we introduce the feature
$f(s_1,\ldots,s_n)$ where $s_i$ is the top-level symbol of $t_i$.
\item[Symbol Features] are various statistics about clause symbols, namely, the
number of occurrences and the maximal depth for each symbol.
\item[Length Features] count the clause length and the numbers of positive
and negative literals.
\item[Conjecture Features] embed information about the conjecture being proved
into the feature vector.  In this way, ENIGMA can provide conjecture-dependent
predictions.
\end{description}

Since there are only finitely many features in any training data, the features
can be serially numbered.
This numbering is fixed for each experiment.
Let $n$ be the number of different features appearing in the training data.
A clause $C$ is translated to a feature vector
$\varphi_C$ whose $i$-th member counts the number of occurrences of the $i$-th 
feature in $C$.
Hence every clause is represented by a sparse numeric vector of length $n$.

With conjecture features, instead of using the vector $\varphi_C$ of length $n$, we use a
vector $(\varphi_C,\varphi_G)$ of length $2n$ where $\varphi_G$ contains the features of
the conjecture $G$.
For a training clause $C$, $G$ corresponds to the conjecture of the proof search
where $C$ was selected as a given clause.
When classifying a clause $C$ during proof search, $G$ corresponds to the conjecture
currently being proved.
When the conjecture consists of several clauses, their vectors are computed
separately and then summed (with the exception of features corresponding to
maxima, such as the maximal symbol depth, where maximum is taken instead).

\subsection{ATP Guidance with Fast Linear Classifiers}
\label{sec:regression}

ENIGMA %
has
so far used simple but fast linear classifiers such as \emph{linear SVM} and \emph{logistic regression} 
efficiently implemented by the LIBLINEAR open source library~\cite{DBLP:journals/jmlr/FanCHWL08}. 
In order to employ them, clause representation by numeric
feature vectors described above in Section~\ref{sec:enigma} is used.
Clausal training data $\T$ are translated to a set of fixed-size labeled
vectors.
Each (typically sparse) vector of length $n$ is labeled either as positive or negative.

The labeled numeric vectors serve as an input to LIBLINEAR which, after the
training, outputs a model $\M$ consisting mainly of a weight vector $w$ of length
$n$. 
The main cost in classifying an arbitrary clause $C$ consists of computation of its feature vector $\varphi_C$ 
and its dot product with the weight vector $\varphi_C\cdot w$. %
ENIGMA assigns to the positively classified clauses a chosen small weight ($1.0$) and a higher weight ($10.0$) to the negatively
classified ones. %
This weight is then used inside E to
guide given clause selection as described in Section~\ref{sec:guidance}.

The training data obtained from the proof runs are typically not balanced with respect
to the number of positive and negative examples.
Usually, there are much more negative examples and the method of
\emph{Accuracy-Balancing Boosting}~\cite{JakubuvU18} was found useful in practice to improve
precision on the positive training data. This is done as follows.
Given training data $\T$ we create a LIBLINEAR classifier $\M$, 
test $\M$ on the training data, and collect the positives mis-classified
by $\M$.
We then repeat (\emph{boost}) the mis-classified positives in the training data, yielding
updated $\T_1$ and an updated classifier $\M_1$.
We iterate this process, and with every iteration, the accuracy on the positive
samples increases, while the accuracy on the negatives typically decreases.
We finish the boosting when the positive accuracy exceeds the negative one.
See~\cite[Sec.2]{JakubuvU18} for details.

\subsection{ATP Guidance with Gradient Boosted Trees}
\label{sec:decision}
Fast linear classifiers together with well-designed features have been
used with good results for a number of tasks in areas such as
NLP~\cite{joulin2017bag}.  However, more advanced learning models have
been recently developed, showing improved performance on a number of
tasks, while maintaining efficiency. One such method is \emph{gradient
  boosted trees} and, in particular, their implementation 
in the XGBoost
library~\cite{DBLP:conf/kdd/ChenG16}. Gradient
  boosted trees are ensembles of decision trees trained by tree boosting.

The format of the training and evaluation data used by XGBoost is
the same as the input used by LIBLINEAR (sparse feature vectors).
Hence, we use practically the
same approach for obtaining the positive and negative training examples, extracting their features, and
clause evaluation during proof runs as described in 
Sections \ref{sec:enigma} and \ref{sec:regression}.
XGBoost, however, does not require the accuracy-balancing boosting. 
This is because XGBoost can deal with unbalanced training data
by setting the ratio of positive and negative examples.\footnote{We use the XGBoost
parameter $\texttt{scale\_pos\_weight}$.}

The model $\M$ produced by XGBoost consists of a set (\emph{ensemble}~\cite{Polikar06}) of decision trees.
The inner nodes of the decision trees consist of conditions on feature values,
while the leafs contain numeric scores.
Given a vector $\varphi_C$ representing a clause $C$, each tree in $\M$ is navigated
to a unique leaf using the values from $\varphi_C$, and the corresponding leaf scores
are aggregated across all trees.
The final score is translated to yield the probability that $\varphi_C$ represents a
positive clause.
When using $\M$ as a weight function in E, the probabilities are turned into binary
classification, assigning weight $1.0$ for probabilities $\ge 0.5$ and
weight $10.0$ otherwise.
Our experiments with scaling of the weight by the probability did not yet yield
improved functionality.

\subsection{Feature Hashing}
\label{sec:hashing}

The vectors representing clauses have so far had length $n$ when $n$
is the total number of features in the training data $\T$ (or $2n$
with conjecture features).  Experiments revealed that both LIBLINEAR
and XGBoost are capable of dealing with vectors up to the length of
$10^5$ with a reasonable performance.  This might be enough for
smaller benchmarks but with the need to train on bigger training data,
we might need to handle much larger feature sets.  In experiments with
the whole translated Mizar Mathematical Library, the feature vector
length can easily grow over $10^6$.  This significantly increases both
the training and the clause evaluation times.  To handle such larger
data sets, we have implemented a simple \emph{hashing} method to
decrease the dimension of the vectors.

Instead of serially numbering all features, %
we represent
each feature $f$ by a unique string and apply a general-purpose string
hashing function to obtain a number $n_f$ within a required range (between 0 and an
adjustable \emph{hash base}).
The value of $f$ is then stored in the feature vector at the position $n_f$.
If different features get mapped to the same
vector index, the corresponding values are summed up.

We use the following hashing function $\mathit{sdbm}$ coming from the open
source SDBM project.
Given a string $s$, the value $h_i$ is computed for every character as follows:
\[
h_i = s_i + (h_{i-1} \ll 6) + (h_{i-1} \ll 16) - h_{i-1}
\]
where $h_0=0$, $s_i$ is the ASCII code of the character at the $i$-th
position, and the operation $\ll$ stands for a bit shift.
The value for the last character is computed with a fixed-size data type (we use
64-bit unsigned integers) and this value modulo the selected hash base is
returned.
We evaluate the effect of the selected hashing function later in
Section~\ref{sec:experiments}.

\section{Neural Architecture for ATP Guidance}
\label{sect:neural_arch}

Although the handcrafted clause features, described in
Section~\ref{sec:enigma}, lead to very good results, they have
naturally several limitations. It is never clear whether the
selected set of features is the best available given the training
data. Moreover, a rich set of features can easily lead to long sparse
vectors and thus using them for large corpora requires the use of
dimensionality reduction techniques (c.f. Section~\ref{sec:hashing}).
Hence selecting the features automatically is a
natural further step.

Among various techniques used to extract features fully automatically,
neural networks (NN) have recently become the most popular thanks to
many successful applications in, e.g., computer vision and natural
language processing. There have been several attempts to use NNs for
guiding ATPs. However, such attempts have so far typically suffered
from a large overhead needed to evaluate the used NN~\cite{LoosISK17},
making them
impractical for actual proving.

A popular approach for representing tree-structured data, like logical
formulae, is based on recursive NNs~\cite{Socher2012}. The basic idea
is that all objects (tree nodes, subterms, subformulas) are represented in a high dimensional vector
space and these representations are subject to learning. Moreover, the
representation of more complex objects is a function of
representations of their arguments. Hence constants, variables, and
atomic predicates are represented as directly learned vectors, called \emph{vector
embeddings}. Assume that all such objects are represented by
$n$-dimensional vectors. For example, constants $a$ and $b$ are
represented by learned vectors $v_a$ and $v_b$, respectively. The
representation of $f(a,b)$ is then produced by a learned function
(NN), say $v_f$, that has as an input two vectors and returns a
vector; hence $v_f(v_a,v_b)\in\R^n$. Moreover, the representation of
$P(f(a, b), a)$ is obtained similarly, because from our point of view
a representation is just a function of arguments. Therefore we have

\smallskip
\begin{tabular}{p{4.75cm}p{10cm}}
  $\displaystyle v_f\colon\underbrace{\R^n\times\dots\times\R^n}_\text{$k$-times}\to\R^n$ & for every $k$-ary, $k\geq 0$, function symbol $f$,\\
  $\displaystyle v_P\colon\underbrace{\R^n\times\dots\times\R^n}_\text{$k$-times}\to\R^n$ & for every $k$-ary, $k\geq 0$, predicate symbol $P$,                                                                                            
\end{tabular}
\smallskip

\noindent in our language. Similarly to ENIGMA, and to make the comparison
better and the model simpler, we also do not distinguish different
variable and Skolem names, e.g., all variables are represented by one
vector. We also replace symbols that appear rarely in our training set
by a representative, e.g., all rare binary functions become the same
binary function. Loosely speaking, we learn a general binary function
this way. Because we treat equality and negation as learned
predicates, we have described how a representation of a literal is
produced.

We could now produce the representation of clauses by assuming that
disjunction is a binary connective, however, we instead use a more
direct approach and we treat clauses directly as sequences of
literals. Recurrent neural networks (RNN) are commonly used to process
arbitrary sequences of vectors. Hence we train an RNN, called $\Cl$,
that consumes the representations of literals in a clause and produces
the representation of the clause,
$\Cl\colon \R^n\times\dots\times\R^n\to\R^n$.

Given a representation of a clause we could learn a function that says
whether the clause is a good given clause. However, without any
context this may be hard to decide. As in ENIGMA and~\cite{LoosISK17},
we introduce more context
into our setting by using the problem's conjecture. 
The negated conjecture is translated by E into a set of clauses. We 
combine the vector representations of these clauses by another RNN, called
$\Conj$ and defined by $\Conj\colon \R^n\times\dots\times\R^n\to\R^n$.

Now we know how to represent a conjecture, by a vector, and a given
clause, by a vector. Hence we can define a function that combines them
into a decision, called $\Fin$ and defined by
$\Fin\colon\R^n\times\R^n\to\R^2$. Note that binary classifications
are commonly represented by functions into $\R^2$, see
Section~\ref{sec:neur-guidance}.

Although all the representations have been vectors in $\R^n$, this is
an unnecessary restriction. It suffices if the objects of the same
type are represented by vectors of the same length. For example, we
have experimented with $\Conj$ where outputs are shorter (and inputs
to $\Fin$ are changed accordingly) with the aim to decrease
overfitting to a particular problem.

\subsection{Neural Model Parameters}
\label{sec:neur-model-param}

The above mentioned neural model can be implemented in many
ways. Although we have not performed an extensive grid search over
various variants, we can discuss some of them shortly. The basic
parameter is the dimension $n$ of the vectors. We have tried various
models with $n\in\{8,16,32,64,128\}$. The functions used for $v_f$ and
$v_P$ can be simple linear transformations (tensors), or more complex
combinations of linear and nonlinear layers. An example of a frequently used nonlinearity is the rectified
linear unit (ReLU), defined by $\max(0,x)$.\footnote{Due to various numerical
  problems with deep recursive networks we have obtained better
  results with ReLU6, defined by $\min(\max(0,x),6)$, or $\tanh$.}

For $\Cl$ and $\Conj$ we use (multi-layer) long short-term memory
(LSTM) RNNs \cite{articleLSTM}. We have tried to restrict the output vector of $\Conj$ to
$m={n\over 2}$ or $m={n\over 4}$ to prevent overfitting with
inconclusive results. The $\Fin$ component is a sequence of
alternating linear and nonlinear layers (ReLU), where the last two
linear layers are $\R^{n+m}\to\R^{n\over 2}$ and
$\R^{n\over 2}\to\R^2$.

\subsection{ATP Guidance with Pytorch}
\label{sec:neur-guidance}

We have created our neural model using the Pytorch library
and integrated it with E using the library's C++ API.\footnote{\url{https://pytorch.org/cppdocs/}}
This API allows to load a previously trained model saved to a file in a special TorchScript format.
We use a separate file for each of the neural parts described above. %
This includes computing of the vector embeddings of terms, literals, and clauses, as well as 
the conjecture embedding $\Conj$ summarizing the %
conjecture clauses into one vector,
and finally the part $\Fin$, which classifies clauses into those 
deemed useful for proving the given conjecture and the rest. 

We have created a new clause weight function in E called TorchEval which interfaces these parts
and can be used for evaluating clauses based on the neural model. One of the key features of the interface,
which is important for ensuring reasonable evaluation speed, is \emph{caching} of the embeddings of terms
and literals. Whenever the evaluation encounters a term or a literal which 
was evaluated before, its embedding is simply retrieved from the memory in constant time instead of being computed 
from the embeddings of its subterms recursively. We use the fact that terms in E are perfectly shared 
and thus a pointer to a particular term can be used as a key for retrieving the corresponding embedding.
Note that this pervasive caching is possible thanks to our choice of recursive neural networks (that match our symbolic data) and it would not work with naive use of other neural models such as convolutional or recurrent networks without their further modifications.

The clause evaluation part of the model returns two real outputs $x_0$ and $x_1$, 
which can be turned into a probability that the given clause will be useful using the sigmoid (logistic) function:
\begin{equation} \label{eq:probability}
p = \frac{1}{1+e^{(x_0-x_1)}}.
\end{equation}
However, for classification, i.e. for a yes-no answer, we can just compare the two numbers 
and ``say yes'' whenever 
\begin{equation} \label{eq:model_said_yes}
x_0 < x_1.
\end{equation}
After experimenting with other schemes that did not perform so well,\footnote{%
For instance, using the probability \eqref{eq:probability} for a more fine-grained order 
on clauses dictated by the neural model.}
we made TorchEval return $1.0$ whenever condition \eqref{eq:model_said_yes} is satisfied
and $10.0$ otherwise. This is in accord with the standard convention employed by E 
that clauses with smaller weight should be preferred and also corresponds to the ENIGMA approach. Moreover,
E implicitly uses an ever increasing clause id as a tie breaker,
so among the clauses within the same class, both TorchEval and ENIGMA behave as FIFO.

Another performance improvement was obtained by forcing Pytorch to use
just a single core when evaluating the model in E.  The default
Pytorch setting was causing degradation of performance on machines
with many cores, probably by assuming by default that multi-threading
will speed up frequent numeric operations such as matrix
multiplication. It seems that in our case, the overhead for
multi-threading at this point may be higher than the gain.

\section{Experimental Evaluation}
\label{sec:experiments}

We experimentally evaluate the three learning-based ATP guidance methods on the MPTP2078
benchmark~\cite{abs-1108-3446}. MPTP2078 contains 2078 problems coming from the MPTP translation~\cite{Urban06} of the
Mizar Mathematical Library (MML) \cite{BancerekBGKMNPU15} to FOL.
The consistent use of symbol names across the MPTP corpus is crucial for our
symbol-based learning methods.
We evaluate ATP performance with a good-performing baseline E strategy, denoted
$\S$, which was previously optimized~\cite{JakubuvU17} on Mizar problems (see
Appendix~\ref{sec:app} for details).

Section~\ref{sec:training} provides details on model training and
the hyperparameters used, and analyzes the most important features used by the tree model.
The model based on linear regression (Section~\ref{sec:regression}) is denoted
$\Mlin$, the model based on decision trees (Section~\ref{sec:decision}) is denoted
$\Mxgb$, and the neural model (Section~\ref{sect:neural_arch}) is denoted
$\Mnn$.
Sections~\ref{sec:atps} and \ref{sec:clauses} evaluate the performance of the models both by 
standard machine learning metrics and by plugging them into the ATPs. 
Section~\ref{sec:hasheval} evaluates the effect of the feature
hashing described in Section~\ref{sec:hashing}.

All experiments were run on a server with 36 hyperthreading Intel(R)
Xeon(R) Gold 6140 CPU @ 2.30GHz cores, with 755 GB of memory available
in total. Each problem is always assigned one core.
For training of the neural models we used NVIDIA GeForce GTX 1080 Ti GPUs. As described above, neither GPU nor multi-threading
is, however, employed when using the trained models for clause evaluation inside the ATP.

\subsection{Model Training, Hyperparameters and Feature Analysis}
\label{sec:training}

We evaluate the baseline strategy $\S$ on all the 2078 benchmark
problems with a fixed CPU time limit of 10 seconds per problem.  This
yields 1086 solved problems and provides the training data for the learning methods as
described in Section~\ref{sec:guidance}.  For $\Mlin$ and $\Mxgb$, the
training data are translated to feature vectors (see
Section~\ref{sec:hand}) which are then fed to the learner.
For $\Mnn$ we use the training data directly without any
feature extraction.

\subsubsection{Training Data and Training of Linear and Tree Models:}

The training data consists of around 222000 training samples (21000 positives and 201000
negatives) with almost 32000 different ENIGMA features.
This means that the training vectors for $\Mlin$ and $\Mxgb$ have dimension close to 64000, and so
has the output weight vector of $\Mlin$.
For $\Mxgb$, we re-use the parameters that performed well in the 
ATPBoost~\cite{PiotrowskiU18} and rlCoP~\cite{KaliszykUMO18} systems 
and produce models with 200 decision trees, each with maximal depth 9.
The resulting models---both linear and boosted trees---are about 1MB large. 
The training time for $\Mlin$ is around 8 minutes (five iterations of
accuracy-balancing boosting), and approximately 5 minutes for $\Mxgb$. 
Both of them are measured on a single CPU core.
During the boosting of $\Mlin$, the positive samples are extended from 21k to 110k
by repeating the mis-classified vectors.

\subsubsection{Learned Tree Features:}
The boosted tree model $\Mxgb$ allows computing statistics of the most
frequently used features. This is an interesting aspect that goes in
the direction of \emph{explainable} AI. The most important features
can be analyzed by ATP developers and compared with the ideas used in
standard clause evaluation heuristics.  There are 200 trees in $\Mxgb$
with 20215 decision nodes in total.  These decision nodes refer to
only 3198 features out of the total 32000.  The most frequently used
feature is the clause length, used 3051 times, followed by the
conjecture length, used 893 times, and by the numbers of the positive
and negative literals in the clauses and conjectures. In a crude way,
the machine learning here seems to confirm the importance assigned to
these basic metrics by ATP researchers.  The set of top ten features
additionally contains three symbol counts (including ``$\in$'' and
``$\subseteq$'') and a vertical feature corresponding to a variable
occurring under negated set membership $\in$ (like in
``$x\not\in\cdot$'' or ``$\cdot\not\in x$'').  This seems plausible,
since the Mizar library and thus MPTP2078 are based on set theory where 
membership and inclusion are the key concepts.

\subsubsection{Neural Training and Final Neural Parameters:}
We try to improve the training of $\Mnn$ by randomly changing the
order of clauses in conjectures, literals in clauses, and terms in
equalities. If after these transformations a negative example, a pair
$(C, G)$, is equivalent to a positive example, we remove the negative
one from the training set. This way we reduce the number of negative
examples to 198k. We train our model in batches\footnote{Moreover, we
  concentrate the examples with the same $G$ into the same batch to
  reduce the training time, because the representation of $G$ has to
  be recomputed for every batch.} of size 128 and use the negative
log-likelihood as a loss function (the learning rate is $10^{-3}$),
where we apply log-softmax on the output of $\Fin$. We weight positive
examples more to simulate a balanced training set. All symbols of the
same type and arity that have less than 10 occurrences in the training
set are represented by one symbol. We set the vector dimension to be
$n=64$ for the neural model $\Mnn$ and we set the output of $\Conj$ to
be $m=16$. All the functions representing function symbols and
predicates are composed of a linear layer and ReLU6. $\Fin$ is set to
be a sequence of linear, ReLU, linear, ReLU, and linear layers. The
training time for $\Mnn$ is around 6 minutes per epoch and the model
was trained for 50 epochs.

\subsection{Evaluation of the Model Performance} 
\label{sec:atps}

\subsubsection{Training Performance of the Models:}
We first evaluate how well the individual models managed
to learn the training data. Due to possible overfitting, this is obviously only used as a heuristic and the main metric is provided by the ultimate ATP evaluation.
Table~\ref{tab:modeleval} shows for each model the \emph{true
  positive} and \emph{true negative} rates (TPR, TNR) on the training data, that
is, the percentage of the positive and negative examples,
classified correctly by each model.\footnote{For $\Mlin$, we show the
  numbers after five iterations of the boosting loop (see
  Section~\ref{sec:regression}).  The values in the first round were
  $\SI{40.81}{\percent}$ for the positive and $\SI{98.62}{\percent}$
  for the negative rate.}
We can see that the highest TPR, also called sensitivity, is achieved by $\Mxgb$
while the highest TNR, also specificity, by $\Mnn$. As expected, the accuracy of the linear model
is lower.
Its main strength seems to come from the relatively high speed of evaluation (see below).

\begin{table}
\caption{True Positive Rate (TPR) and True Negative Rate (TNR) on training data.} %
\label{tab:modeleval}
\vspace{-6mm}
\center
\bgroup
\setlength\tabcolsep{2mm}
\begin{tabular}{c|ccc}
	& $\Mlin$ & $\Mxgb$ & $\Mnn$  \\
\hline
TPR & $\SI{90.54}{\percent}$ & $\SI{99.36}{\percent}$ & $\SI{97.82}{\percent}$\\
TNR & $\SI{83.52}{\percent}$ & $\SI{93.32}{\percent}$ & $\SI{94.69}{\percent}$\\
\end{tabular}
\egroup
\vspace{-6mm}
\end{table}

\subsubsection{ATP Performance of the Models:} 
Table~\ref{tab:reallife} shows the total number of problems solved by
the four methods.  For each learning-based model $\M$, we always
consider the model alone ($\S\odot\M$) and the model combined equally
with $\S$ ($\S\oplus\M$). All methods are using the same time limit,
i.e., 10 seconds. This is our ultimate ``real-life'' evaluation,
confirming that the boosted trees indeed outperform the guidance by
the linear classifier and that the recursive neural network and its
caching implementation is already competitive with these methods in
real time.  The best method $\S\oplus\Mxgb$ solves $\SI{15.7}{\percent}$ more
problems than the original strategy $\S$, and $\SI{3.8}{\percent}$ problems more
than the previously best linear strategy $\S\oplus\Mlin$.\footnote{We have also measured how much $\S$ benefits from increased time limits. It solves 1099 problems in \SI{20}{\second} and 1137 problems in \SI{300}{\second}.}
Table~\ref{tab:reallife} provides also further explanation of
these aggregated numbers.  We show the number of unique solutions
provided by each of the methods and the difference to the original
strategy. 
Table~\ref{tab:greedy} shows how useful are the particular methods when used together. Both the linear and the neural models complement the boosted trees well, while the original strategy is made completely redundant.
\begin{table}
\caption{Number of problems solved (and uniquely solved) by the individual models. $\S+$ and $\S-$ are the additions and missing solutions wrt  $\S$.}
\label{tab:reallife}
\center
\bgroup
\setlength\tabcolsep{1mm}
\begin{tabular}{l|c|cc|cc|cc}
	    & $\S$ & $\S\odot\Mlin$ & $\S\oplus\Mlin$ & $\S\odot\Mxgb$ & $\S\oplus\Mxgb$ & $\S\odot\Mnn$ & $\S\oplus\Mnn$ \\
\midrule
solved &    1086  & 1115 & 1210 & 1231 & \textbf{1256} & 1167 & 1197 \\
unique &  0 &  3 &  7 &  10  &  \textbf{15} &  3 &  2 \\
$\S+$ &  0 & +119 & +138 & +155 & \textbf{+173} & +114 & +119 \\
$\S-$ & 0 &   -90 & -14 & -10 & \textbf{-3} & -33 & -8  \\
\end{tabular}
\egroup
\vspace{-6mm}
\end{table}
\begin{table}
\caption{The greedy sequence---methods sorted by their greedily computed contribution to all the problems solved.}
\label{tab:greedy}
\vspace{-6mm}
\center
\bgroup
\setlength\tabcolsep{1mm}
\begin{tabular}{l|ccccccc}
greedy sequence	    & $\S\oplus\Mxgb$ & $\S\oplus\Mlin$ & $\S\odot\Mnn$ & $\S\odot\Mxgb$ & $\S\odot\Mlin$ & $\S\oplus\Mnn$ & $\S$ \\
\hline 
greedy addition &  1256 & 33 & 13 & 11 & 3 & 2 & 0 \\
greedy total &  1256 &  1289 &  1302 &  1313 &  1316 & 1318 & 1318 \\
\end{tabular}
\egroup
\vspace{-6mm}
\end{table}
%
%
%
%
%
%
%

\subsubsection{Testing Performance of the Models on Newly Solved Problems:}
There are 232 problems newly solved---some of them multiple times---by
the six learning-based methods. To see how the trained models behave
on new data, we again extract positive and negative examples from all
successful proof runs on these problems. This results in around
\num{31000} positive testing examples and around \num{300000} negative testing examples.

Table~\ref{tab:modeleval1} shows again for each of the previously trained models the \emph{true
  positive} and \emph{true negative} rates (TPR, TNR) on these testing data.
The highest TPR is again achieved by $\Mxgb$
and the highest TNR by $\Mnn$. The accuracy of the linear model
is again lower. Both the TPR and TNR testing scores are for all methods 
significantly lower compared to their training counterparts. TPR decreases by about $\SI{15}{\percent}$
and TNR by about $\SI{20}{\percent}$. This likely shows the limits of our current learning and proof-state characterization methods. It also points to the very interesting issue of obtaining many \emph{alternative proofs}~\cite{KuhlweinU12b} and learning from them. It seems that just using learning or reasoning is not sufficient in our AI domain, and feedback loops combining the two multiple times~\cite{US+08-long,PiotrowskiU18} are really necessary for building strong ATP systems.

\begin{table}
\caption{True Positive Rate (TPR) and True Negative Rate (TNR) on testing data from the newly solved 232 problems.} %
\label{tab:modeleval1}
\center
\bgroup
\setlength\tabcolsep{2mm}
\begin{tabular}{c|ccc}
	& $\Mlin$ & $\Mxgb$ & $\Mnn$  \\
\hline
  TPR & $\SI{80.54}{\percent}$ & $\SI{83.35}{\percent}$ & $\SI{82.00}{\percent}$\\
TNR & $\SI{62.28}{\percent}$ & $\SI{72.60}{\percent}$ & $\SI{76.88}{\percent}$\\
\end{tabular}
\egroup
\vspace{-6mm}
\end{table}

\subsection{Speed of Clause Evaluation by the Learned Models}
\label{sec:clauses}

The number of generated clauses reported by E can be used as a rough estimate 
of the amount of work done by the prover. If we compare this statistic for those
runs that timed out---i.e., did not find a proof within the given time limit---we
can use it to estimate the slowdown of the clause processing rate incurred by 
employing a machine learner inside E. (Note that each generated clause needs to be
evaluated before it is inserted on the respective queue.)

Complementarily, the number of processed clauses compared across the problems 
on which all runs succeeded may be seen as an indicator of how well the respective
clause selection guides the search towards a proof (with a perfect guidance,
we only ever process those clauses which constitute a proof).\footnote{This metric is similar
in spirit to \emph{given clause utilization} introduced by Schulz and M\"ohrmann \cite{DBLP:conf/cade/SchulzM16}.}

\begin{table}
\caption{%
The ASRPA and NSRGA ratios. ASRPA are the average ratios (and standard deviations) of the relative number of \emph{processed} clauses 
with respect to $\S$ on problems on which all runs succeeded. 
NSRGA are the average ratios (and standard deviations) of the relative number of \emph{generated} clauses 
with respect to $\S$ on problems on which all runs timed out.
The numbers of problems were \num{898} and \num{681}, respectively.}
\label{tab:genproc}
\vspace{-6mm}
\center
\bgroup
\begin{tabular}{l|c|c|c|c|c|c|c}
	    & $\S$ & $\S\odot\Mlin$& $\S\oplus\Mlin$ & $\S\odot\Mxgb$ & $\S\oplus\Mxgb$ & $\S\odot\Mnn$  &  $\S\oplus\Mnn$ \\
\hline
ASRPA   & $1 \pm 0$ & $2.18 \pm 20.35$ & $0.91 \pm 0.58$ & $0.60 \pm 0.98$ & $0.59 \pm 0.36$ & $0.59 \pm 0.75$ & $0.69 \pm 0.94$ \\
NSRGA  & $1 \pm 0$ & $0.61 \pm \phantom{0}0.52$ & $0.56 \pm 0.35$ & $0.42 \pm 0.38$ & $0.43 \pm 0.35$ & $0.06 \pm 0.08$ & $0.07 \pm 0.09$ \\
\end{tabular}
\egroup
\vspace{-6mm}
\end{table}

Table~\ref{tab:genproc} compares the individual configurations of E
based on the seven evaluated models with respect to these two metrics.
To obtain the shown values, we first normalized the numbers on per
problem basis with respect to the result of the baseline strategy $\S$
and computed an average across all relevant problems.
The comparison of thus obtained All Solved Relative Processed Average
(ASRPA) values shows that, with the exception of $\S\odot\Mlin$ (which
has a very high standard deviation), all other configurations on
average manage to improve over $\S$ and find the corresponding proofs
with fewer iterations of the given clause loop. This indicates better
guidance towards the proof on the selected benchmarks.

The None Solved Relative Generated Average (NSRGA) values represent
the speed of the clause evaluation. It can be seen that while the linear
model is relatively fast (approximately $\SI{60}{\percent}$ of the
speed of $\S$), followed closely by the tree-based model  (around
$\SI{40}{\percent}$), the neural model is more expensive to evaluate
(achieving between $6$ and $\SI{7}{\percent}$ of $\S$).  

We note that
without caching, NSRGA of $\S\oplus\Mnn$ drops from $7.1$ to
$\SI{3.6}{\percent}$ of the speed of $\S$. Thus caching currently helps to
approximately double the speed of the evaluation of clauses with
$\Mnn$.\footnote{Note that more global caching (of e.g.~whole clauses and frequent combinations of literals) across multiple problems may further amortize the cost of the neural evaluation. This is left as future work here. }
 It is interesting and encouraging that despite the neural method being currently about ten times slower than the linear method---and thus generating about ten times
fewer inferences within the $\SI{10}{\second}$ time limit used for the ATP evaluation (Table~\ref{tab:reallife})---the neural model already manages to outperform the linear model in the unassisted setting. I.e., $\S\odot\Mnn$ is already better than $\S\odot\Mlin$, despite the latter being much faster.

\subsection{Evaluation of Feature Hashing}
\label{sec:hasheval}

Finally, we evaluate the feature hashing described in Section~\ref{sec:hashing}. 
We try different hash bases in order to reduce dimensionality of the vectors and to estimate
the influence on the ATP performance.
We evaluate on 6 hash bases from 32k ($2^{15}$), 16k ($2^{14}$), down to 1k ($2^{10}$).
For each hash base, we construct models $\Mlin$ and $\Mxgb$, we compute their
prediction rates, and evaluate their ATP performance.

With the hash base $n$, each feature must fall into one of $n$
\emph{buckets}.
When the number of features is greater than the base---which is our case as we
intend to use hashing for dimensionality reduction---collisions are inevitable.
When using hash base of 32000 (ca $2^{15}$) there are almost as many
hashing buckets as there are features in the training data (31675).
Out of these features, ca 12000 features are hashed without a collision and 12000 buckets
are unoccupied.
This yields a $\SI{40}{\percent}$ probability of a collision.
With lower bases, the collisions are evenly distributed.

\begin{table}[t]
\caption{Effect of feature hashing on prediction rates and ATP performance.}
\label{tab:hashing}
\vspace{-6mm}
\center
\bgroup
\setlength\tabcolsep{2mm}
\begin{tabular}{c|c|cccccccc}
\multicolumn{2}{c|}{model $\backslash$ hash size}
          & \textit{without} & 32k & 16k    & 8k     & 4k     & 2k     & 1k    \\
\hline
&&&&&&&&&\\[-3mm]
\multirow{4}{*}{$\Mlin$} &
   TPR [\%]    & 90.54 &
      89.32 & 88.27 & 89.92 & 82.08 & \textbf{91.08} & 83.68
\\
   & TNR [\%]    & 83.52 & 82.40 & \textbf{86.01} & 83.02 & 81.50 & 76.04 & 77.53
\\
   & $\S\odot\M$  &  \textbf{1115} & 1106 & 1072 & 1078 & 1076 & 1028 & 938
\\
   & $\S\oplus\M$  & \textbf{1210} & 1218 & 1189 & 1202 & 1189 & 1183 & 1119
\\\hline
&&&&&&&&&\\[-3mm]
\multirow{4}{*}{$\Mxgb$} &
   TPR [\%]    & 99.36    & 99.38  & 99.38  & 99.51  & 99.62   &  99.65
   & \textbf{99.69}
\\ & 
   TNR [\%]   & 93.32    & 93.54  & 93.29  & 93.69  & 93.90    &  94.53
   & \textbf{94.88}
\\
   & $\S\odot\M$  &
   1231 & 1231 & \textbf{1233} & 1232 & 1223 & 1227 & 1215 
\\
   & $\S\oplus\M$  & \textbf{1256} 
        & 1244 & 1244 & \textbf{1256} & 1245 & 1236 & 1232
\end{tabular}
\egroup
\vspace{-6mm}
\end{table}

Lower hash bases lead to larger loss of information, %
hence
decreased performance can be expected.
On the other hand, dimensionality reduction sometimes leads to better generalization (less overfitting of the learners). 
Also, the
evaluation in the ATP can be done more efficiently in a lower
dimension, thus giving the ATP the chance to process more clauses.
The prediction rates and ATP performance for models with and without hashing are
presented in Table~\ref{tab:hashing}.
We compute the true positive (TPR) and negative (TNR) rates as in
Section~\ref{sec:training}, and we again evaluate E's performance 
based on the strategy $\S$ in the two ways ($\odot$ and $\oplus$) as in
Section~\ref{sec:atps}.
The best value in each row is highlighted.
Both models perform comparably to the version without hashing 
even when the vector dimension is reduced to $\SI{25}{\percent}$.
With reduction to 1000 (32x), the models still provide a decent improvement
over the baseline strategy $\S$.
The $\Mxgb$ model deals with the reduction slightly better.

Interestingly, the classification accuracy of the models (again, measured only on the training data) seems to increase with the decrease of hash base
(especially for $\Mxgb$).
However, with this increased accuracy, the ATP performance mildly decreases.  
This could be caused by the more frequent collisions and thus learning on data that have been made less precise.

\section{Conclusions and Future Work}
\label{sec:conclude}
We have described an efficient implementation of gradient-boosted and
recursive neural guidance in E, extending the ENIGMA framework.  The
tree-based models improve on the previously used linear classifier,
while the neural methods have been for the first time shown
practically competitive and useful, by using extensive caching
corresponding to the term sharing implemented in E.  While this is
clearly not the last word in this area, we believe that this is the
first practically convincing application of gradient-boosted and
neural clause guidance in saturation-style automated theorem provers.

There are a number of future directions. For example, research in
better proof state characterization of saturation-style systems has
been started
recently~\cite{DBLP:conf/itp/GoertzelJ0U18,ProofWatch_Meets_ENIGMA_First}
and it is likely that evolving vectorial representations of the proof
state will further contribute to the quality of the learning-based
guidance. Our recursive neural model is just one of many, and a number
of related and combined models can be experimented with.

\label{sect:bib}
\bibliographystyle{plain}

\bibliography{aitp,ate11,nn}

\appendix

\section{Strategy $\S$ from Experiments in Section~\ref{sec:experiments}}
\label{sec:app}

The following E strategy has been used to undertake the experimental evaluation
in Section~\ref{sec:experiments}.  The given clause selection strategy
(heuristic) is defined using parameter ``\verb+-H+''.

\begin{verbatim}
--definitional-cnf=24 --split-aggressive --simul-paramod -tKBO6 -c1 -F1
-Ginvfreq -winvfreqrank --forward-context-sr --destructive-er-aggressive 
--destructive-er --prefer-initial-clauses -WSelectMaxLComplexAvoidPosPred
-H'(1*ConjectureTermPrefixWeight(DeferSOS,1,3,0.1,5,0,0.1,1,4),
    1*ConjectureTermPrefixWeight(DeferSOS,1,3,0.5,100,0,0.2,0.2,4),
    1*Refinedweight(ConstPrio,4,300,4,4,0.7),
    1*RelevanceLevelWeight2(PreferProcessed,0,1,2,1,1,1,200,200,2.5, 
                                                           9999.9,9999.9),
    1*StaggeredWeight(DeferSOS,1),
    1*SymbolTypeweight(DeferSOS,18,7,-2,5,9999.9,2,1.5),
    2*Clauseweight(ConstPrio,20,9999,4),
    2*ConjectureSymbolWeight(DeferSOS,9999,20,50,-1,50,3,3,0.5),
    2*StaggeredWeight(DeferSOS,2))'
\end{verbatim}

\end{document}